\documentclass[final,3p,times,twocolumn]{elsarticle}
\usepackage{amsmath}
\usepackage{amssymb}
\usepackage{booktabs}
\usepackage{diagbox}
\usepackage{color}
\usepackage{multirow}
\usepackage{bm}
\usepackage{lineno,hyperref}
\usepackage{threeparttable}
\usepackage{wrapfig}
\usepackage{pifont}
\modulolinenumbers[5]
\journal{arXiv}

\begin{document}

\begin{frontmatter}



    \title{F-SE-LSTM: A Time Series Anomaly Detection Method with Frequency Domain Information}


    \author[address1]{Yi-Xiang Lu}
    \author[address2]{Xiao-Bo Jin\corref{corresponding}}
    \cortext[corresponding]{Corresponding author}
    \ead{xiaobo.jin@xjtlu.edu.cn}
    \author[address3]{Jian Chen}
    \author[address1]{Dong-Jie Liu}
    \author[address1]{Guang-Gang Geng}

    \affiliation[address1]{
        organization={College of Cyber Security},
        addressline={Jinan University},
        city={Guangzhou},
        postcode={510632},
        state={Guangdong},
        country={China}}

    \affiliation[address2]{
        organization={Intelligent Science Department},
        addressline={Xi'an Jiaotong-Liverpool University},
        city={Suzhou},
        postcode={215123},
        state={Jiangsu},
        country={China}}

    \affiliation[address3]{
        organization={Institute of Industrial Internet and Internet of Things},
        addressline={China Academy of Information and Communications Technology},
        city={Beijing},
        postcode={100191},
        country={China}}

    \begin{abstract}
        With the development of society, time series anomaly detection plays an important role in network and IoT services.
        However, most existing anomaly detection methods directly analyze time series in the time domain and cannot distinguish some relatively hidden anomaly sequences.
        We attempt to analyze the impact of frequency on time series from a frequency domain perspective, thus proposing a new time series anomaly detection method called F-SE-LSTM.
        This method utilizes two sliding windows and fast Fourier transform (FFT) to construct a frequency matrix.
        Simultaneously, Squeeze-and-Excitation Networks (SENet) and Long Short-Term Memory (LSTM) are employed to extract frequency-related features within and between periods.
        Through comparative experiments on multiple datasets such as Yahoo Webscope S5 and Numenta Anomaly Benchmark, the results demonstrate that the frequency matrix constructed by F-SE-LSTM exhibits better discriminative ability than ordinary time domain and frequency domain data.
        Furthermore, F-SE-LSTM outperforms existing state-of-the-art deep learning anomaly detection methods in terms of anomaly detection capability and execution efficiency.
    \end{abstract}



    % \begin{highlights}
    % \end{highlights}


\end{frontmatter}


\section{Introduction}

\begin{figure}[htbp]
    \centering
    \includegraphics[width=0.45\textwidth]{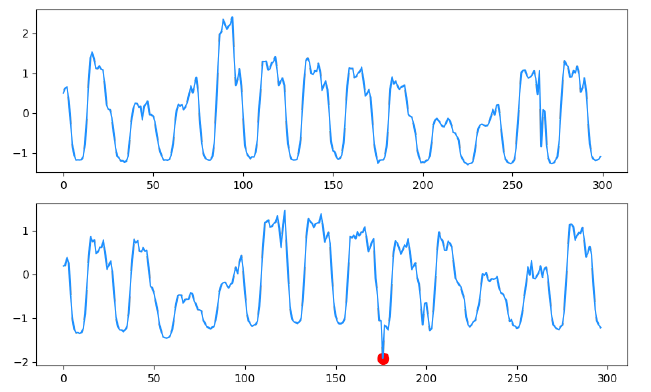}
    \caption{Two network traffic time series data show similar trends in time.}
    \label{time_series}
\end{figure}

With the development of society, time series data closely related to people are generated in large quantities in various fields, such as:
network traffic data generated when surfing the Internet \cite{ahmed2016survey},
transaction data generated by online transactions with others \cite{seyedhossein2010mining},
bioelectric signal data displayed on the electrocardiogram \cite{sivaraks2015robust},
flow data generated by traffic monitoring equipment \cite{zhang2022elastic}
and sensor data generated by Internet of Things (IoT) devices \cite{cook2019anomaly}, etc.
Fig. \ref{time_series} shows two time series of network traffic data, which exhibit similar trends over time, and thus have the same intrinsic regularity.
However, in the event of an attack on the server, network intrusion, traffic congestion, or network equipment failure that poses a threat to network security, the inherent patterns in the time series of this network traffic will be disrupted, leading to the emergence of abnormal points or anomalous data trends.
Therefore, anomaly detection of time series plays an important role in reducing economic losses, maintaining network security and social stability.

The detection of outliers in time series data \cite{fox1972outliers} has been a subject of research since the last century.
After years of development, various methods based on statistical models suitable for anomaly detection have been explored in the field of time series analysis.
These methods include autoregressive (AR) \cite{kay1982robust}, moving average (MA) \cite{ohsaki2009analysis}, autoregressive moving average (ARMA) \cite{akay1991application} and autoregressive integrated moving average (ARIMA) \cite{nelson1998time}.
With the widespread adoption of machine learning and deep learning in artificial intelligence in recent years, many researchers have proposed novel time series anomaly detection methods based on these models.
These include techniques such as K-Means \cite{rebbapragada2009finding}, local outlier factor \cite{oehmcke2015event}, one-class support vector machines \cite{scholkopf1999support}, convolutional neural network (CNN) \cite{munir2018deepant} and recurrent neural network (RNN) \cite{su2019robust}, among others.

\begin{figure}[htbp]
    \centering
    \includegraphics[width=0.45\textwidth]{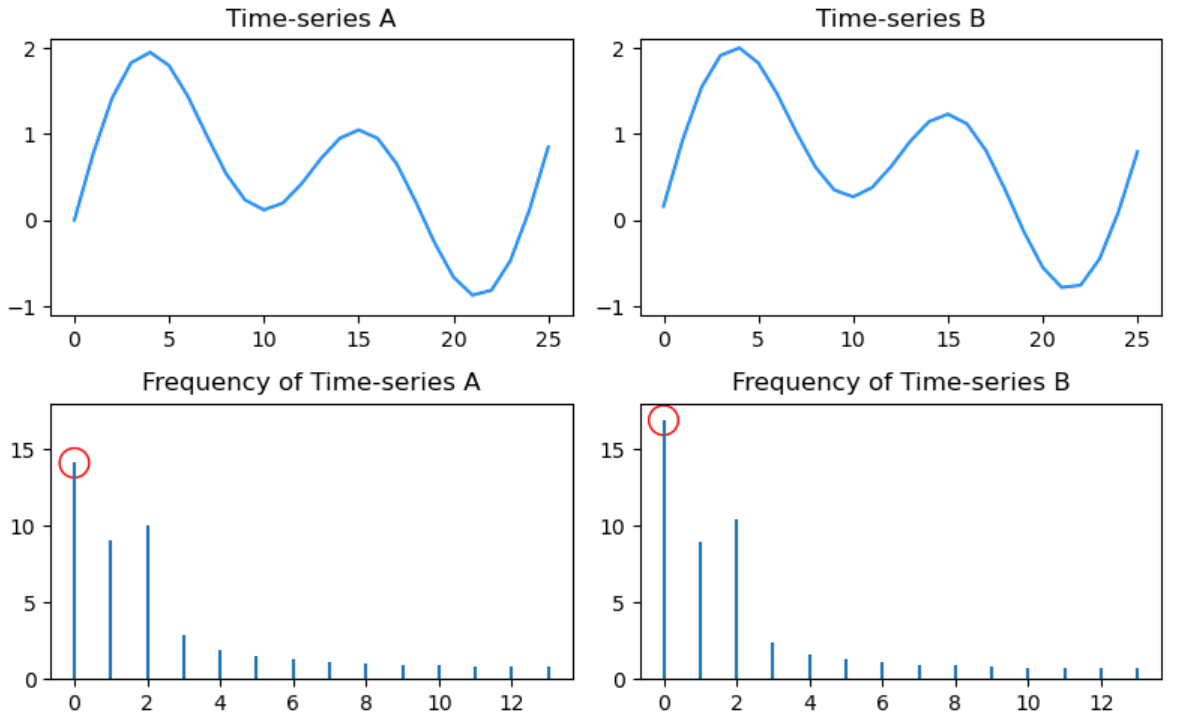}
    \caption{Two sequences A and B that are very similar in the time domain have significant differences in the frequency domain at specific frequencies.}
    \label{time_freq}
\end{figure}

Since the time series statistical model is specially designed for time series data, the principle and structure of the model are relatively simple, and the applicable data types are limited.
Generally, time domain analysis can only be performed based on the original data of the time series.
In the time domain, although this method can quickly capture some obvious outliers, it is difficult to observe the differences in the time domain for some relatively similar sequences.
Therefore, we need to adjust the perspective and try to observe the time series from other domains.
As shown in Fig. \ref{time_freq}, sequence A and sequence B are very similar in time domain, and it is difficult to distinguish the difference between them, but in the frequency domain, the two sequences will have different amplitudes at certain frequencies, where these differences help to distinguish two time series that are very similar in the time domain.

Although machine learning and deep learning models are more complex, they can be used on a wide range of data types, including raw time series data and its derived features, such as statistical feature data and frequency domain data.
In addition, the parameters in the model can also be adaptively adjusted according to different data, so that the model can fully capture the characteristics of the data.
Therefore, time series anomaly detection methods based on machine learning or deep learning models can detect some relatively hidden anomalies, thereby obtaining more accurate anomaly detection results.

Only a few methods currently use frequency domain information in deep learning, and these methods only convert time series data directly into frequency domain data \cite{rahimi2020deep}\cite{soleimani2023effectiveness}.
However, time series data typically changes over time and has temporal correlation in the time domain.
These methods directly convert time series into frequency data, which destroys the original temporal correlation information of the data and does not consider the frequency correlation in the same time period and between different time periods.

In this paper, we propose an anomaly detection method for time series based on frequency domain analysis and deep learning models, which combines fast Fourier transform (FFT) \cite{cochran1967fast}, Squeeze-and-Excitation networks (SENet) \cite{hu2018squeeze} and Long Short-Term Memory (LSTM) \cite{hochreiter1997long}, called F-SE-LSTM.
FFT is a frequency domain transformation method that can efficiently convert time series with time domain data into frequency domain data.
Compared to other frequency domain transformation methods, FFT is more efficient and has a wide range of applications in signal processing, communication, and other fields.
SENet is a channel attention network based on CNN, and LSTM is a variant of RNN, which can avoid gradient disappearance and explosion during training for long sequences.
On the anomaly detection problem of time series, we first use FFT to convert time domain data of different time periods of time series into frequency domain data and construct a frequency matrix,
and then adopt SENet and LSTM to extract the frequency-related information within and between time periods,
and finally apply a deep neural network (DNN) to output binary classification results for anomaly detection.
We run experiments on some real datasets from Yahoo Webscope S5 and Numenta Anomaly Benchmark to verify the anomaly detection ability of the proposed method.
Experimental results on F1, recall and precision show that the proposed method shows excellent performance on these datasets, and its anomaly detection ability outperforms existing state-of-the-art methods.

The main contributions of this paper include the following three points:
\begin{itemize}
    \item Proposal of a method utilizing FFT and sliding window to transform time series data into a frequency matrix, which encompasses frequency-related features and frequency-based time-related features. This method offers valuable insights for the frequency domain analysis of time series data.
    \item Integration of SENet and LSTM to construct time series anomaly detection models. This framework enables the extraction of frequency-related information within the same period and between different periods, enhancing the characterization of anomalies in time series data.
    \item Experimental results demonstrate that our proposed method, F-SE-LSTM, outperforms existing anomaly detection methods in terms of detection capability and execution efficiency. This contribution presents a practical solution to the problem time series anomaly detection.
\end{itemize}

The remainder of this paper is organized as follows.
In Section 2, we review various time series anomaly detection methods, analyze their strengths and weaknesses, and then lead to the proposed anomaly detection method.
In Section 3, we describe our proposed F-SE-LSTM method in detail.
In Section 4, we describe the experimental procedure and provide a comprehensive comparison of our method with other methods.
Finally, we conclude this paper with a final section.

\section{Related work}

With the development of science and technology, machine learning and deep learning models play an increasingly important role in many applications, especially in the field of artificial intelligence and have achieved good application results.
In the field of time series, machine learning and deep learning models can also better fit the characteristics of time series.
Therefore, many researchers have gradually applied machine learning and deep learning models to time series anomaly detection.
Up to now, many time series anomaly detection methods based on machine learning and deep learning models have been explored.
We summarize these methods into methods based on machine learning models, methods based on convolutional neural networks, methods based on recurrent neural networks, and methods based on various deep learning models according to the different algorithms used.
These methods are summarized below and their advantages and disadvantages are analyzed.

\textbf{Methods based on machine learning models}.
This method mainly uses machine learning models and some feature extraction algorithms to construct a time series anomaly detection method.
For example, Anton et al. \cite{anton2019anomaly} preprocess temporal data of network traffic time series using principal component analysis and detect anomalies using SVM and random forest classifiers to address network intrusion.
Experiments have verified that the above two machine learning models have good effects in industrial Internet applications on two network attack datasets in industrial environments.
To make up for the lack of a single model, Zhou et al. \cite{zhou2021anomaly} fused k-nearest neighbor, symbolic aggregation approximation, and Markov model into one fusion model, and proposed an anomaly detection method called KSM.
This method mainly performs time domain analysis on the time series data in order to extract the features of amplitude variation and sequence shape, which can detect anomalies caused by shape and amplitude variation more accurately.
This method has higher data anomaly identification ability than most single methods.
In addition, some researchers have adopted the method of frequency domain spatial embedding to convert time series time domain data into frequency domain data, construct frequency domain features and then use machine learning models for detection.
For example, He and Pi \cite{he2016anomaly} used short-time Fourier transform (STFT) to transform time series time domain data into frequency domain data, and then implemented early warning and detection of helicopter rotor failure using a one-class classification method based on support vector data description (SVDD).
Experiments show that the method can clearly detect the damage of helicopter rotor on the vibration data of a real helicopter.

\textbf{Methods based on convolutional neural networks}.
This method mainly uses convolutional neural network to construct time series anomaly detection method.
For example, Hwang et al. \cite{hwang2020unsupervised} proposed a malicious traffic detection mechanism D-PACK based on network traffic time domain data by combining CNN and Autoencoder.
This mechanism can only detect the first few bytes of the first few packets of each flow, which greatly reduces the data input, thereby improving the detection speed, and its accuracy is not much different from full detection, which can be better applied to industry.
Wen and Keyes \cite{wen2019time} proposed a convolutional U-Net time series segmentation approach based on time series raw data for anomaly detection, which is the first time to use a CNN-based deep network for time series segmentation and shows better anomaly detection ability in time series.
In order to solve the network intrusion problem, Ullah and Mahmoud \cite{ullah2021design} constructed network traffic into one-dimensional, two-dimensional and three-dimensional data, and constructed CNN anomaly detection models including four convolution and pooling layers.
The experimental results show that the three CNN anomaly detection models are superior to other existing deep anomaly detection models.
Neupane et al. \cite{neupane2022temporal} built a Temporal Convolution Networks (TCN) based on CNN to detect abnormal vehicle sensor data.
Experimental results show that the model can achieve high anomaly detection accuracy.
Djamal et al. \cite{djamal2019detection} used FFT to convert time domain data of the EEG time series into frequency domain data, and then used a CNN model to detect abnormal sequences.
Experiments show that FFT can greatly improve the accuracy of detecting time domain data, while CNN can extract appropriate frequency features, further improving the ability of anomaly detection.
Rahimi et al. \cite{rahimi2020deep} used FFT to extract acoustic emission signal features for detecting fuel tank leaks in cars, and used one-dimensional convolutional neural networks for detection.
Experiments show that using FFT for feature extraction is more effective and can achieve better detection results compared to wavelet transform and time-domain statistical features.

\textbf{Methods based on recurrent neural networks}.
This method mainly adopts recurrent neural network or its variants to construct time series anomaly detection method.
For example, Zhang and Zou \cite{zhang2018time} used an LSTM neural network model to predict sequences from light curve data and output prediction errors for anomaly detection.
Experimental results show that the model has good application effects in light curve time series prediction and anomaly detection.
To address the problem of anomaly detection in time series, Malhotra et al. \cite{malhotra2015long} used stacked LSTM networks to forecast time series and detect anomalous sequence through prediction errors.
The model has produced promising results on many different types of real-world datasets, and also verified that the LSTM-based model outperforms the RNN-based model in time series anomaly detection.
For some unpredictable time series, they proposed an LSTM-based Encoder-Decoder model to reconstruct time series and detect abnormal sequences through the reconstruction error \cite{malhotra2016lstm}.
Experimental results show that the model can detect some unpredictable abnormal sequences, reflecting the robustness of the model.
Wang et al. \cite{wang2022improved} improved LSTM to predict time series data in rail transit systems and detect abnormal data based on prediction errors.
A large number of experiments have been carried out in the real subway operation environment to verify the effectiveness of the proposed scheme, and at the same time, it has better performance than the existing anomaly detection methods.
Kieu et al. \cite{kieu2019outlier} proposed two ensemble frameworks for time series anomaly detection based on recurrent autoencoders, one weight-independent and the other weight-shared.
Both frameworks use sparsely connected recurrent neural networks to build autoencoders, which can reduce the impact of autoencoders on outlier overfitting, thereby improving the overall detection performance.
In detecting abnormal traffic in Border Gateway Protocol (BGP), Cheng et al. \cite{cheng2016ms} proposed a multi-scale LSTM model that can learn the long dependencies in temporal patterns of time domain data.
Experimental results on real BGP datasets show that the model is about 10\% more accurate than LSTM and common machine learning models.
To detect various violent activities, Park et al. \cite{park2020punch} used FFT to convert the accelerometer and gyroscope time series data into frequency domain data to analyze the frequency characteristics, and then used the LSTM model to classify punches, which can effectively detect violent activities and quickly respond to violent crimes.

\textbf{Methods based on various deep learning models}.
This method uses a variety of existing neural networks to construct a time series anomaly detection method.
For example, Kim and Cho \cite{kim2018web} adopted the C-LSTM model \cite{zhou2015c} proposed by Zhou et al. to detect anomalous network traffic.
The model combines CNN and LSTM, which can simultaneously extract spatial and temporal features of time series data, and has better anomaly detection ability than other machine learning and deep learning models.
Yin et al. \cite{yin2020anomaly} stacked multiple time series into a time series matrix.
At the same time, referring to the C-LSTM model structure, they combined CNN and LSTM-based autoencoder to construct a C-LSTM-AE model suitable for time series matrices.
The model can well extract the time-related features generated by the time series matrix, and achieve better detection performance than C-LSTM on the network traffic dataset.

Among these time series anomaly detection methods, although the machine learning method can effectively detect the anomaly of time series data, this method simply regards the time series as a vector in a multidimensional space without considering the relationship between each data point.
Therefore, it is difficult to extract the deep features of time series.
The deep learning method can focus on the association between data points and can better extract the characteristics of time series.
For example, the convolutional neural network can extract the spatial correlation features of the time series, and the recurrent neural network can extract the temporal correlation features.
In addition, a small number of researchers use the frequency domain space embedding method to convert time series time domain data into frequency domain vectors, construct features that cannot be expressed in time domain space, and combine deep learning models to more fully capture the characteristics of time series.
However, most of the existing frequency domain methods only use frequency-related features without considering the correlation frequency features and time, and do not fully play their corresponding roles according to the characteristics of various deep learning models.
Therefore, these frequency domain methods also have a lot of room for improvement in the anomaly detection ability of time series.
To this end, we use FFT to convert time series of different time periods into frequency bands to construct frequency information containing time order, and use appropriate deep learning models to fully extract frequency-related features containing the time information, thereby constructing a more accurate and effective time series anomaly detection method.

\section{Methodology}

In order to detect hidden abnormal data more accurately, and improve the anomaly detection ability of time series, a time series anomaly detection method F-SE-LSTM based on frequency domain is proposed,
which includes the following steps: (1) Construct frequency matrix data by FFT; (2) Extract features of frequency matrix by using SENet and LSTM; (3) Output anomaly detection results by DNN.
Below we first describe the steps to create the frequency matrix, then describe the construction of the anomaly detection model, and finally describe the parameter settings of the proposed method.

\subsection{Creation of frequency matrix}
\begin{figure*}[htbp]
    \centering
    \includegraphics[width=\textwidth]{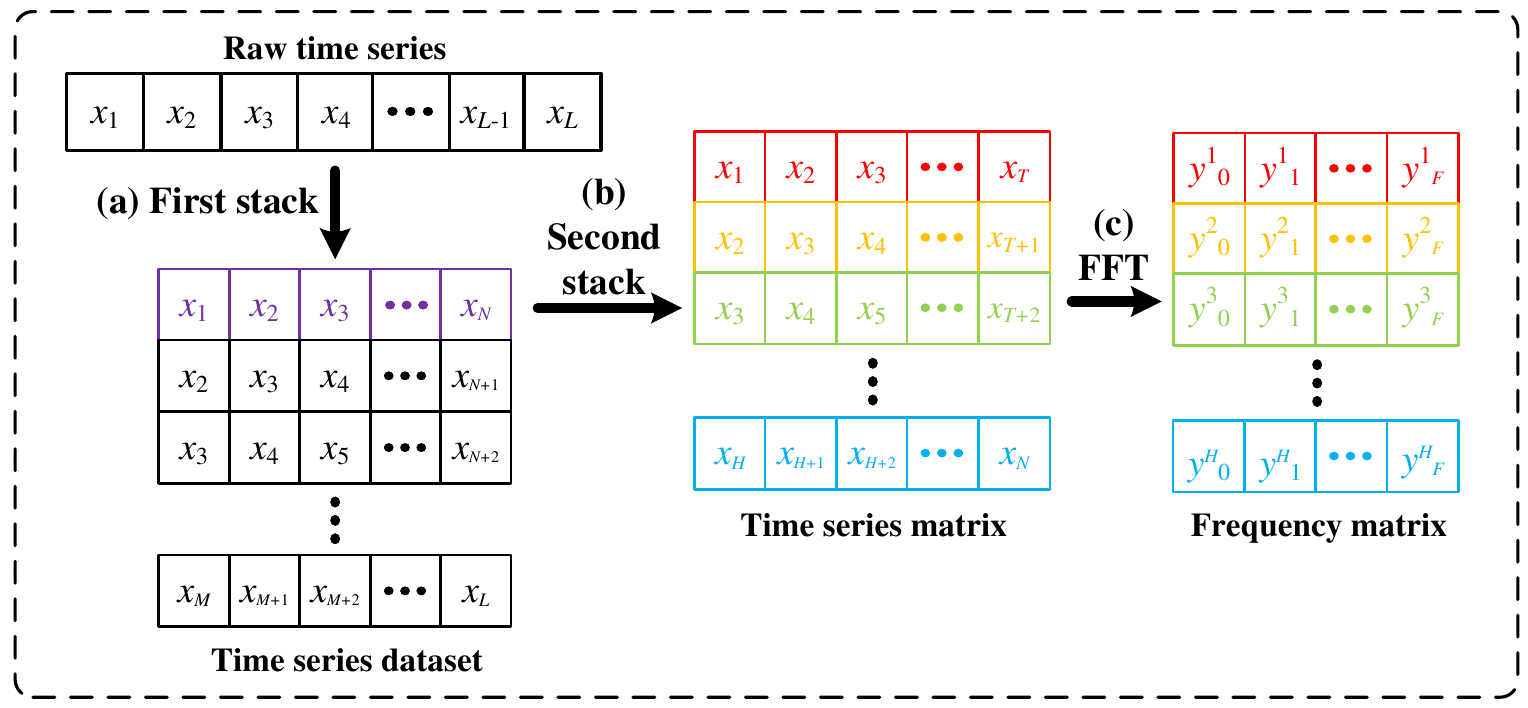}
    \caption{Frequency matrix creation steps.}
    \label{frequency_matrix}
\end{figure*}

Consider a time series $\bm{x}=(x_1,x_2,...,x_L)$ of length $L$, as shown in Fig. \ref{frequency_matrix}(a), the time series $\bm{x}$ is divided into $M$ sequences of length $N$ through the sliding window, namely
\begin{equation}
    \bm{D} =
    \begin{bmatrix}
        x_1    & x_2     & \cdots & x_N     \\
        x_2    & x_3     & \cdots & x_{N+1} \\
        \vdots & \vdots  & \vdots & \vdots  \\
        x_M    & x_{M+1} & \cdots & x_L
    \end{bmatrix},
\end{equation}
where $M=L-N+1$, and each sequence represents each sample.
Each sample in the dataset that contains an outlier is considered an abnormal sample, and a sample that does not contain an outlier is considered a normal sample.
In order to make full use of the temporal correlation information of the samples and distinguish the normal samples and abnormal samples as accurately as possible, as shown in Fig. \ref{frequency_matrix}(b), we continue to divide the samples of length $N$ into $H$ subsequences of length $T$ through another sliding window, and finally each sample is converted into a time series matrix:
\begin{equation}
    \bm{S^t} =
    \begin{bmatrix}
        x_1    & x_2     & \cdots & x_T     \\
        x_2    & x_3     & \cdots & x_{T+1} \\
        \vdots & \vdots  & \vdots & \vdots  \\
        x_H    & x_{H+1} & \cdots & x_N
    \end{bmatrix},
\end{equation}
where $H=N-T+1$.

% \begin{equation}
%     f(t) = a_0+\sum_{n=1}^{\infty}\big(a_n\cos(\frac{2\pi n}{T}t)+b_n\sin(\frac{2\pi n}{T}t)\big)
%     \label{fs1}
% \end{equation}

To find the abnormal samples that are difficult to identify in the time domain, we transform the time series matrix into other domains to construct relevant features.
The Fourier series is a special triangular series that can approximately represent any periodic function that meets the Dirichlet Conditions, which provides the basis for transformations to other domains.
In general, we discuss the Fourier series in the complex domain by introducing Euler's formula, as in the following equation:
\begin{equation}
    f(t) = \sum_{n=-\infty}^{\infty}c_n e^{i\frac{2\pi n}{T}t},
    \label{fs}
\end{equation}
where $t$ is time, $n$ is frequency, and $i$ is an imaginary unit.
In order to transform the function from time domain space to frequency domain space and expand it to non periodic function, Fourier transform (FT) was developed.
FT can directly realize the transformation of the signal from the time domain space to the frequency domain space by the following equation:
\begin{equation}
    F(\omega)=\int_{-\infty}^{\infty}f(t)e^{-i\omega t}dt,
    \label{ft}
\end{equation}
where $\omega$ is angular frequency, $F(\omega)$ is frequency domain function, which plays a key role in the field of signal processing.
% \begin{equation}
%     c_n=\frac{1}{T}\int_0^Tf(t)e^{-i\frac{2\pi n}{T}t}dt, \quad n = 0,1,2,\cdots,T-1,
%     \label{ft}
% \end{equation}

Due to the inability to obtain continuous signals in practical applications, signals can only be sampled at certain time intervals.
To describe the frequency domain of discrete sequence data, we introduce the discrete Fourier Transform (DFT):
% \begin{equation}
%     X(n) = \sum_{t=0}^{T-1}e^{-i\frac{2\pi n}{T}t}\cdot x(t), \quad n = 0,1,2,\cdots,T-1,
%     \label{dft}
% \end{equation}
\begin{equation}
    X(n) = \sum_{t=0}^{T-1}e^{-in\omega t}\cdot x(t), \quad n = 0,1,2,\cdots,T-1,
    \label{dft}
\end{equation}
where the input $\{x(t)\}$ is the time domain data, and the output $\{X(n)\}$ is the frequency domain data of the complex plane.
It is necessary to take the modulus of the complex number $X(n)$ to obtain the frequency amplitude $y(n)$ for facilitating subsequent calculation:
\begin{equation}
    y(n) = \lvert X(n) \rvert, \quad n = 0,1,2,\cdots,T-1.
    \label{mod}
\end{equation}

When calculating Eqn. (\ref{dft}), FFT can divide the time $t$ in the equation into odd and even numbers and add them separately.
The idea of divide and conquer reduces the number of operations and greatly improves the operation efficiency of DFT operations.
In order to construct the relevant features in the frequency domain space, as shown in Fig. \ref{frequency_matrix}(c), each sequence in the time series matrix is converted into frequency band amplitudes by using FFT, and the frequency matrix is further constructed.
Since the amplitudes of frequency $n$ and $T-n$ are equal according to the Eqn. (\ref{dft}) and (\ref{mod}), only $\lfloor T/2 \rfloor$ frequencies in the $X(n)$ contain useful information, so the frequency matrix is:
\begin{equation}
    \bm{S^f} =
    \begin{bmatrix}
        y^1_0  & y^1_1  & \cdots & y^1_F  \\
        y^2_0  & y^2_1  & \cdots & y^2_F  \\
        \vdots & \vdots & \vdots & \vdots \\
        y^H_0  & y^H_1  & \cdots & y^H_F
    \end{bmatrix},
\end{equation}
where $F=\lfloor T/2 \rfloor$.

\subsection{Deep anomaly detection model}

\begin{figure*}[htbp]
    \centering
    \includegraphics[width=\textwidth]{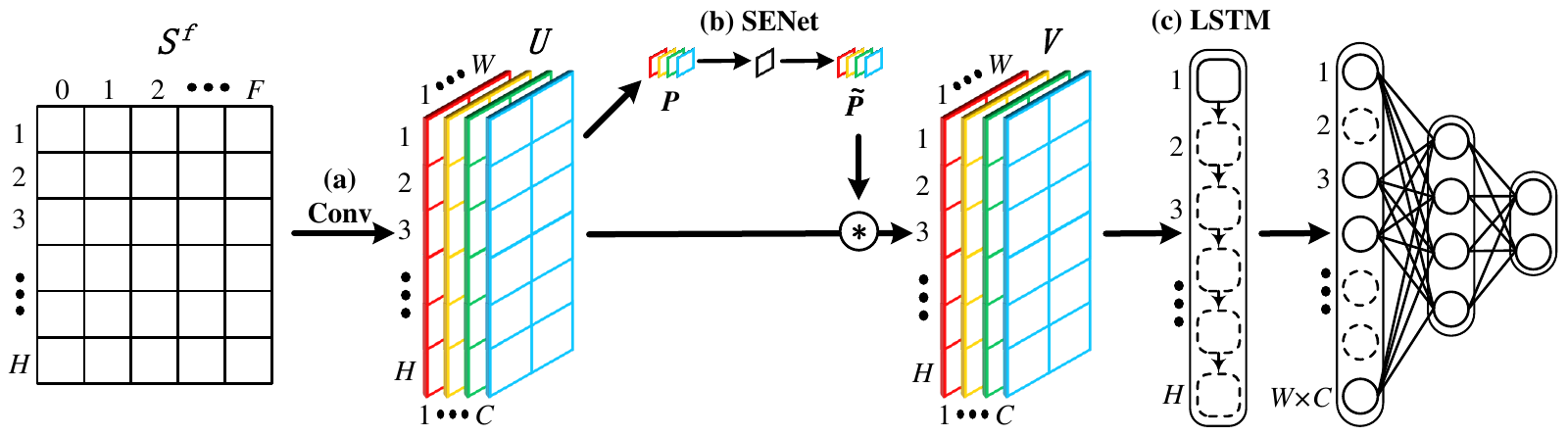}
    \caption{
        Anomaly detection model:
        (a) Frequency matrix is convoluted;
        (b) SENet is used to extract the dependencies between frequencies in the same time period;
        (c) LSTM is adopted to extract the dependencies between frequencies in different time periods.
    }
    \label{detection_model}
\end{figure*}

Here, we introduce a deep learning model combining SENet and LSTM for anomaly detection.
SENet is a CNN with channel attention mechanism, which can improve accuracy by modeling the correlation between feature channels and strengthening important features.
It can extract not only the interdependence between image pixels, but also the interdependence between channels, and has higher feature extraction capabilities than CNN.
Therefore, SENet is suitable for extracting the dependencies between frequencies within a certain time period.
On the other hand, LSTM, as a variant of RNN \cite{zaremba2014recurrent}, can well model the temporal context dependencies of long and short sequences, and thus can model the dependencies between frequencies of different periods.
Combining the above two neural network models can allow the detection algorithms to give full play to their respective advantages in time series anomaly detection, so as to build a more robust anomaly detection model.

For the constructed frequency matrix $\bm{S^f}\in\mathbb{R}^{H\times(F+1)}$, as shown in Fig. \ref{detection_model}(a), SENet first convolves the adjacent frequencies in each time period to obtain $\bm{U}\in\mathbb{R}^{H\times W\times C}$ for extracting the dependencies between adjacent frequencies, where $W$ is the frequency dimension size after convolution, and $C$ is the channel dimension size.
In Fig. \ref{detection_model}(b), the matrix $\bm{U_C}\in\mathbb{R}^{H\times W}$ of each channel in $\bm{U}$ is squeezed to obtain $\bm{P}\in\mathbb{R}^{1\times1\times C}$ for extracting the dependency between channels by the following equation:
\begin{equation}
    \bm{P_C}=\frac{1}{H\times W}\sum_{i=1}^H\sum_{j=1}^W\bm{U_C}(i,j),
    \label{se1}
\end{equation}
where $\bm{P_C}\in\mathbb{R}^{1\times1}$ is the matrix of the channel dimension of the tensor $\bm{P}$.
Then according the compress and excitation mechanism, the correlation coefficient tensor $\bm{\widetilde{P}}\in\mathbb{R}^{1\times1\times C}$ is obtained by convolving the channels of $\bm{P}$.
Finally, each channel vector $\bm{U_{H\times W}}\in\mathbb{R}^C$ in $\bm{U}$ is operated with $\bm{\widetilde{P}}$ such as:
\begin{equation}
    \bm{V_{H\times W}}=\tanh(\bm{U_{H\times W}}*\bm{\widetilde{P}}),
    \label{se2}
\end{equation}
where $\tanh$ is the activation function, the symbol $*$ is the Hadamard product, and the $\bm{V_{H\times W}}\in\mathbb{R}^{C}$ is the channel vector in tensor $\bm{V}\in\mathbb{R}^{H\times W\times C}$ which contains frequency and channel related information.

In the next step, we extract the frequency-related feature between different time periods through LSTM.
The $H$ output matrices $\bm{V_H}\in\mathbb{R}^{W\times C}$ on the time dimension are taken as $H$ time steps and each matrix is spliced into a one-dimensional vector $z_t$ of size $W\times C$, and then input into the LSTM neural network.
LSTM contains LSTM cells that process the input data of each time step in a recurrent fashion.
LSTM cells include forget gates, input gates, output gates, cell states and hidden states.
The forget gate, input gate and output gate in Eqn. (\ref{fgate}), (\ref{igate}) and (\ref{ogate}) can operate the input data $z_t$ and the hidden state $h_{t-1}$ to obtain the corresponding gating coefficients, where $\sigma$ is the sigmoid activation function, and $\bm{W}$ and $\bm{b}$ are weights and biases.
Then, as shown in Eqn. (\ref{celltran}), (\ref{cellstate}) and (\ref{hiddenstate}), the forget gate can choose to forget the unimportant information in the previous cell state, and the input gate can retain the important information of the input data, so as to obtain the new cell state $c_t$.
Finally, the output gate determines the important information that the current cell state can output, and use the new hidden state $h_t$ as the output of LSTM.

\begin{eqnarray}
    f_t & = & \sigma(\bm{W^f}[h_{t-1},z_t] + \bm{b^f})
    \label{fgate} \\
    i_t & = & \sigma(\bm{W^i}[h_{t-1},z_t] + \bm{b^i})
    \label{igate}  \\
    o_t & = & \sigma(\bm{W^o}[h_{t-1},z_t] + \bm{b^o})
    \label{ogate} \\
    \tilde{c}_t & = & \tanh(\bm{W^c}[h_{t-1},z_t] + \bm{b^c})
    \label{celltran} \\
    c_t & = & f_t * c_{t-1} + i_t * \tilde{c}_t
    \label{cellstate} \\
    h_t & = & o_t * \tanh(c_t)
    \label{hiddenstate}
\end{eqnarray}

In order to output time series anomaly detection results, it is necessary to use DNN to reduce the dimensionality of the one-dimensional vector output by LSTM.
At the same time, DNN introduces dropout to prevent overfitting, that is, randomly discards a certain proportion of neurons during each training, and finally outputs the results of anomaly detection binary classification.

\subsection{Parameter configurations}

This section will describe the parameters of the constructed frequency matrix and the parameters of the model structure to ensure that the proposed method can be better adapted to our problem.

When constructing sample sequences, we follow the settings in the literature and set the sample sequence length $N$ to 60.
For the construction of the frequency matrix, we set the size of the sliding window $T$ to 30 to balance time information and frequency information.
Then the number of subsequences $H$ is 31, the maximum frequency $F$ is 15, and the shape of the final frequency matrix $\bm{S^f}$ is $31\times16$.

\begin{table}[htbp]
    \centering
    \caption{Parameters settings of SENet in F-SE-LSTM}
    % \vspace{2px}
    \scalebox{0.75}{
        \begin{tabular}{l|c|c|c}
            \hline
            \multicolumn{4}{c}{\textbf{SENet}}                                                                                                  \\
            \hline
            \textbf{Type}               & \textbf{\#Filter}                                            & \textbf{Kernel size} & \textbf{Stride} \\
            \hline
            \textbf{Conv2d \ding{172}}  & 16                                                           & 1$\times$4           & 1$\times$4      \\
            \textbf{AvgPood2d}          & -                                                            & 31$\times$4          & 31$\times$4     \\
            \textbf{Conv2d}             & 4                                                            & 1$\times$1           & 1$\times$1      \\
            \textbf{ReLU}               & -                                                            & -                    & -               \\
            \textbf{Conv2d}             & 16                                                           & 1$\times$1           & 1$\times$1      \\
            \textbf{Sigmoid \ding{173}} & -                                                            & -                    & -               \\
            \hline
            \textbf{Hadamard product}   & \multicolumn{3}{c}{the outputs of \ding{172} and \ding{173}}                                          \\
            \hline
            \textbf{Tanh}               & -                                                            & -                    & -               \\
            \hline
        \end{tabular}
    }
    \label{senet_param}
\end{table}

\begin{table}[htbp]
    \centering
    \caption{Parameters settings of LSTM and DNN in F-SE-LSTM}
    % \vspace{2px}
    \scalebox{0.75}{
        \begin{tabular}{l|c|c}
            \hline
            \multicolumn{3}{c}{\textbf{LSTM \& DNN}}                            \\
            \hline
            \textbf{Type}         & \textbf{Hidden size} & \textbf{Output size} \\
            \hline
            \textbf{LSTM}         & 64                   & 64                   \\
            \textbf{Dropout(0.2)} & -                    & -                    \\
            \textbf{Linear}       & -                    & 32                   \\
            \textbf{Tanh}         & -                    & -                    \\
            \textbf{Linear}       & -                    & 2                    \\
            \hline
        \end{tabular}
    }
    \label{lstm_dnn_param}
\end{table}

For the anomaly detection model, the parameters of SENet, LSTM and DNN are shown in the Table \ref{senet_param} and \ref{lstm_dnn_param}.
In SENet, 16 convolution kernels of size and stride $1\times4$ output a tensor $\bm{U}$ of shape $31\times4\times16$.
Next, the tensor $\bm{U}$ is first input to the average pooling layer of size and stride $31\times4$, and the output tensor $\bm{P}$ of shape $1\times1\times16$.
Then, compress the channel dimension by convolution and ReLU activation function to output a tensor of shape $1\times1\times4$, and output a tensor $\bm{\widetilde{P}}$ of shape $1\times1\times16$ by convolution and sigmoid activation function.
Finally, the Hadamard product and tanh activation function are used for $\bm{U}$ and $\bm{\widetilde{P}}$ according to Eqn. (\ref{se2}), and output the tensor $\bm{V}$ of shape $31\times4\times16$.
In LSTM and DNN, the tensor $\bm{V}$ is stretched into a matrix of shape $31\times64$ as the input of LSTM, and the vector of size 64 is output to DNN to obtain the final binary classification result.

\section{Experiment}

In this section, we compare our method with other methods on publicly available datasets.
A hardware environment consisting of Intel (R) Xeon (R) Gold 5218 CPU, A100-PCIE-40GB GPU and 256 GB RAM was used.
In terms of software, we mainly use the Sklearn machine learning framework and the PyTorch deep learning framework to build time series anomaly detection methods.

\subsection{Datasets}

We select 4 time series real datasets related to network traffic and Internet devices for experiments: the A1 dataset in Yahoo Webscope S5 \cite{laptevlabeled} and the AWS, Known, Traffic datasets in Numenta Anomaly Benchmark \cite{numenta}.
The A1 dataset contains 67 sequences of web traffic from Yahoo, each containing 1000 observations.
The AWS dataset contains 17 sequences, including the CPU usage rate of Amazon services, the number of network input bytes, and the number of disk read bytes, and most of the sequences have more than 4,000 observations.
The Known dataset contains 7 sequences, including temperature sensor data and CPU utilization, with thousands to tens of thousands of observations.
The Traffic dataset contains 7 series of vehicle sensor occupancy, speed and travel data, each containing 1000 to 3000 observations.

\subsection{Data preprocessing}

Since the time series value range of each dataset is different, we normalized each original sequence to a sequence with a mean of 0 and a standard deviation of 1, and then used the method described in the previous section to create a sample with a sequence length of 60.
Random stratified sampling is used to divide the dataset into training set, validation set and test set according to the ratio of $6:2:2$, where the proportion of abnormal samples in the three sets is roughly equal.
The total number of samples and the proportion of abnormal samples in each dataset are shown in Table \ref{dataset}.

\begin{table}[htbp]
    \centering
    \caption{Samples size and abnormal proportion of datasets}
    \vspace{2px}
    \begin{tabular}{l|c|c}
        \hline
        \textbf{Datasets} & \textbf{Number} & \textbf{Proportion} \\
        \hline
        \textbf{A1}       & 90913           & 9.41\%              \\
        \textbf{AWS}      & 66737           & 2.79\%              \\
        \textbf{Know}     & 69148           & 1.66\%              \\
        \textbf{Traffic}  & 15251           & 5.35\%              \\
        \hline
    \end{tabular}
    \label{dataset}
\end{table}

\subsection{Evaluation metrics}

In order to fully verify the advantages and disadvantages of the proposed method, multiple evaluation metrics are used: accuracy, precision, recall and F1, as shown in:
\begin{eqnarray}
    Accuracy & = &\frac{TP+TN}{TP+FP+FN+TN},
    \label{a} \\
    Precision& = &\frac{TP}{TP+FP},
    \label{p} \\
    Recall& = &\frac{TP}{TP+FN},
    \label{r} \\
    F1& =&2\times \frac{Precision\times Recall}{Precision+Recall},
    \label{f1}
\end{eqnarray}
where abnormal samples are marked as true positive (TP) when the prediction is correct,
normal samples are marked as false positive (FP) when the prediction is wrong,
abnormal samples are marked as false negative (FN) when the prediction is wrong,
and normal samples are marked as true negative (TN) when the prediction is correct.

Generally, accuracy is the proportion of all samples correctly predicted.
Precision is the correct proportion of predicted abnormal samples, and the larger the value, the lower the abnormal false alarm rate.
Recall is the proportion of correct prediction in abnormal samples, and the larger the value, the lower the abnormal missed detection rate.
F1 combines precision and recall, while the higher the value, the better the comprehensive detection effect.
In order to measure the comprehensive performance of each method, F1 is used as the main evaluation metric.

\subsection{Results and analysis}

\begin{table*}[htbp]
    \centering
    \caption{Comparison of frequency domain features and time domain features on multiple datasets and multiple models}
    \vspace{2px}
    \scalebox{0.9}{
        \begin{tabular}{l|c|c|c|c|c|c|c|c}
            \hline
            \multirow{2}*{\textbf{Models}} & \multicolumn{2}{c|}{\textbf{A1}} & \multicolumn{2}{c|}{\textbf{AWS}} & \multicolumn{2}{c|}{\textbf{Known}} & \multicolumn{2}{c}{\textbf{Traffic}}                                                                   \\
            \cline{2-9}
                                           & \textbf{Time}                    & \textbf{Freq.}                    & \textbf{Time}                       & \textbf{Freq.}                       & \textbf{Time} & \textbf{Freq.} & \textbf{Time} & \textbf{Freq.} \\
            \hline
            \textbf{kNN}                   & 0.6010                           & 0.9106                            & 0.3414                              & 0.9149                               & 0.5790        & 0.8039         & 0.4901        & 0.9479         \\
            \textbf{LR}                    & 0.1063                           & 0.5101                            & 0.0317                              & 0.4216                               & 0.0029        & 0.2382         & 0.0122        & 0.5925         \\
            \textbf{SVM}                   & 0.5688                           & 0.6760                            & 0.3448                              & 0.7216                               & 0.2734        & 0.4951         & 0.5427        & 0.8370         \\
            \textbf{DT}                    & 0.5917                           & 0.7463                            & 0.3417                              & 0.8063                               & 0.3798        & 0.5875         & 0.4555        & 0.8157         \\
            \textbf{RF}                    & 0.6907                           & 0.8183                            & 0.3560                              & 0.8765                               & 0.4342        & 0.6296         & 0.5458        & 0.9003         \\
            \hline
        \end{tabular}
    }
    % \begin{tablenotes}
    %     \footnotesize
    %     \item 
    % \end{tablenotes}
    \label{ml}
\end{table*}

We set up the following sets of experiments to fully verify the effectiveness of each step in the proposed method, and the experimental codes\footnote{Code URL: https://github.com/lyx199504/f-se-lstm-time-series} are uploaded to GitHub.
In order to ensure the fairness of the experiment, the machine learning models used in the experiments are all set to default parameters, and the hyperparameters learning rate, batch size, epochs and random state of the deep learning model are set to 0.001, 512, 500 and 1, respectively.
In the training of the deep learning model, we take the sum of the F1 scores of each epoch and the 5 epochs before and after (in total 11 epochs) on the validation set as the validation score.
Then the number of epochs with the highest validation score is applied for testing on the test set.

\subsubsection{Effectiveness of frequency domain features}
We first used FFT to convert sequence samples of length 60 into frequency bands, and then used various machine learning models to conduct experiments with time domain data and frequency domain data, including k-nearest neighbor (kNN), logical regression (LR), support vector machine (SVM), decision tree (DT) and random forest (RF).
From the results in Table \ref{ml}, it can be seen that the F1 score of frequency domain data is higher than that of time domain data on various models and multiple datasets, which shows that frequency domain data is more discriminative than time domain data in anomaly detection.
% , including statistical models, linear models, tree models and ensemble models.

\begin{figure*}[htbp]
    \centering
    \includegraphics[width=\textwidth]{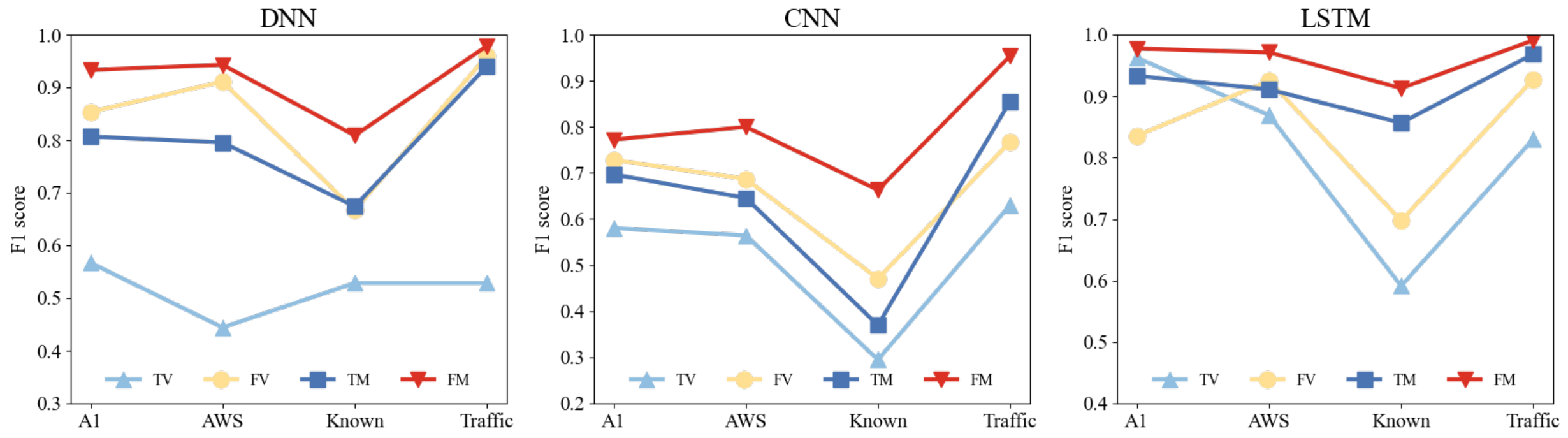}
    \caption{Influence of frequency matrices on various deep learning algorithms and datasets.}
    \label{ex2}
\end{figure*}

\subsubsection{Effectiveness of frequency matrix}
Since the deep learning model can directly process matrix data, we used three commonly used neural networks, such as DNN, CNN and LSTM, to detect time series vectors (Fig. \ref{frequency_matrix}(a)), frequency vectors (vectors undergo FFT transformation), time series matrices (Fig. \ref{frequency_matrix}(b)) and frequency matrices (Fig. \ref{frequency_matrix}(c)).
As shown in Fig. \ref{ex2}, each line represents the F1 score of the model on different datasets, where TV, FV, TM and FM represent the input data as time series vectors, frequency vectors, time series matrices and frequency matrices.
It can be observed that the F1 score of frequency data of the same dimension is higher than that of time series data, and the F1 score of high-dimensional data of the same data type is higher than that of low dimensional data, indicating that the frequency matrix as input data can fully capture the time series features, which can achieve better results in anomaly detection.

\subsubsection{Parameter exploration of frequency matrix}
To further verify the impact of sliding window size $T$ on the performance of various deep learning algorithms, we set $T$ to 10, 20, 30, 40, and 50 respectively, and conducted comparative experiments on DNN, CNN and LSTM.
The experimental results are shown in the Table \ref{dl}.
When $T$ is 30, the total number of the highest F1 scores is far more than the total number of other values of $T$, and for each model, the value has advantages in at least half of the datasets.
In general, this value balances the number of frequencies and time periods in the sequence samples, making the time information and frequency information extracted by the model more sufficient.

\begin{table*}[htbp]
    \centering
    \caption{F1 scores of various deep learning algorithms under different sliding windows}
    % \vspace{2px}
    \scalebox{0.9}{
        \begin{tabular}{l|c|c|c|c|c|c|c|c|c|c|c|c}
            \hline
            \multirow{2}*{$T$} & \multicolumn{3}{c|}{\textbf{A1}} & \multicolumn{3}{c|}{\textbf{AWS}} & \multicolumn{3}{c|}{\textbf{Known}} & \multicolumn{3}{c}{\textbf{Traffic}}                                                                                                                                                 \\
            \cline{2-13}
                               & \textbf{DNN}                     & \textbf{CNN}                      & \textbf{LSTM}                       & \textbf{DNN}                         & \textbf{CNN}    & \textbf{LSTM}   & \textbf{DNN}    & \textbf{CNN}    & \textbf{LSTM}   & \textbf{DNN}    & \textbf{CNN}    & \textbf{LSTM}   \\
            \hline
            10                 & 0.7845                           & 0.6193                            & 0.9720                              & 0.7779                               & 0.7742          & 0.9554          & 0.6340          & 0.5062          & 0.8488          & 0.9527          & 0.9363          & 0.9876          \\
            20                 & 0.8958                           & 0.7851                            & 0.9727                              & 0.9131                               & 0.7512          & 0.9686          & 0.7877          & 0.4570          & 0.9005          & 0.9653          & 0.9196          & 0.9877          \\
            30                 & 0.9331                           & 0.7723                            & \textbf{0.9774}                     & \textbf{0.9427}                      & \textbf{0.8000} & \textbf{0.9712} & \textbf{0.8087} & 0.6631          & 0.9129          & 0.9785          & \textbf{0.9536} & \textbf{0.9908} \\
            40                 & 0.9205                           & \textbf{0.8009}                   & 0.9716                              & 0.9365                               & 0.7035          & 0.9699          & 0.7857          & \textbf{0.8038} & 0.9057          & \textbf{0.9877} & 0.9494          & 0.9877          \\
            50                 & \textbf{0.9434}                  & 0.7753                            & 0.9709                              & \textbf{0.9427}                      & 0.7236          & 0.9671          & 0.7938          & 0.5647          & \textbf{0.9138} & 0.9750          & 0.8715          & 0.9907          \\
            \hline
        \end{tabular}
    }
    \label{dl}
\end{table*}

\subsubsection{Effectiveness of proposed model}
In order to verify that the model constructed by the method in this paper can fully and effectively extract the features of the frequency matrix, we performed two experiments.
For the first experiment we used the constructed frequency matrix as input data and trained using LSTM+DNN, CNN+LSTM+DNN and SENet+LSTM+DNN (the proposed model).
In the second experiment, we used the time matrix and frequency matrix as input data respectively, and used the proposed model for training.

\begin{figure}[htbp]
    \centering
    \includegraphics[width=0.45\textwidth]{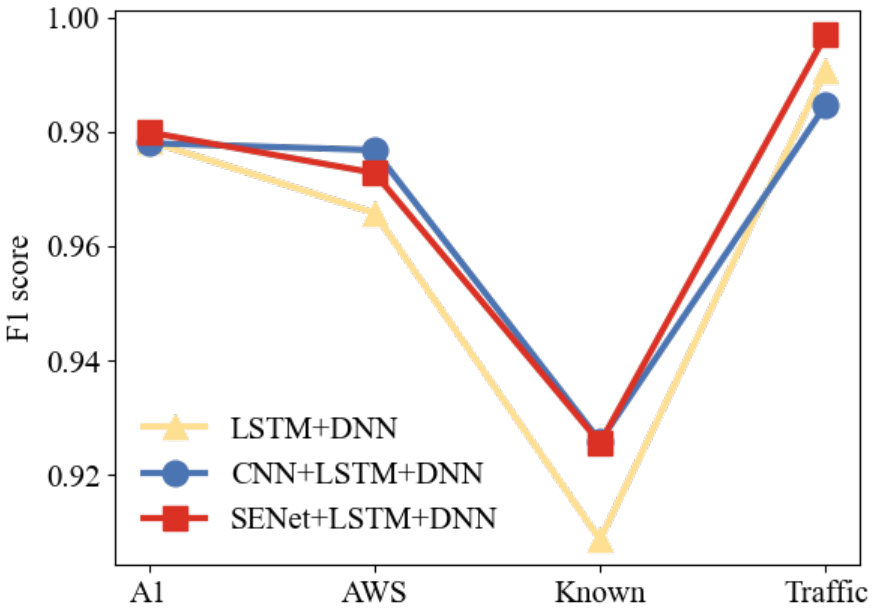}
    \caption{Comparison of different model structures on different datasets.}
    \label{ex4}
\end{figure}

The first experimental result is shown in Fig. \ref{ex4}.
The proposed model outperforms LSTM+DNN in all four datasets and achieves the highest F1 scores on two datasets.
Although CNN+LSTM+DNN also achieved the highest F1 score on two datasets, it is lower than LSTM+DNN on the other two datasets.
This shows that SENet performs more stable than CNN in time series anomaly detection.
The second experimental result is shown in Table \ref{e4_t}.
The F1 score obtained by the proposed model with frequency matrix as input is better than that with time matrix as input on all four datasets.

Therefore, the idea of using SENet to extract channel-related features is valid, and the proposed model is also more suitable for extracting features of frequency matrices.

\begin{table*}[htbp]
    \centering
    \caption{Comparison of two methods on F1 scores in each dataset}
    \vspace{2px}
    % \scalebox{0.8}{
    \begin{tabular}{l|cccc}
        \toprule
        \textbf{Methods}                 & \textbf{A1}     & \textbf{AWS}    & \textbf{Known}  & \textbf{Traffic} \\
        \midrule
        SENet+LSTM+DNN[Time Matrix]      & 0.9627          & 0.9316          & 0.8868          & 0.9753           \\
        SENet+LSTM+DNN[Frequency Matrix] & \textbf{0.9799} & \textbf{0.9728} & \textbf{0.9256} & \textbf{0.9969}  \\
        \bottomrule
    \end{tabular}
    % }
    % \begin{tablenotes}
    %     \footnotesize
    %     \item The time column represents how long the anomaly detection method was trained on the A1 dataset.
    % \end{tablenotes}
    \label{e4_t}
\end{table*}

\begin{figure*}[htbp]
    \centering
    \includegraphics[width=\textwidth]{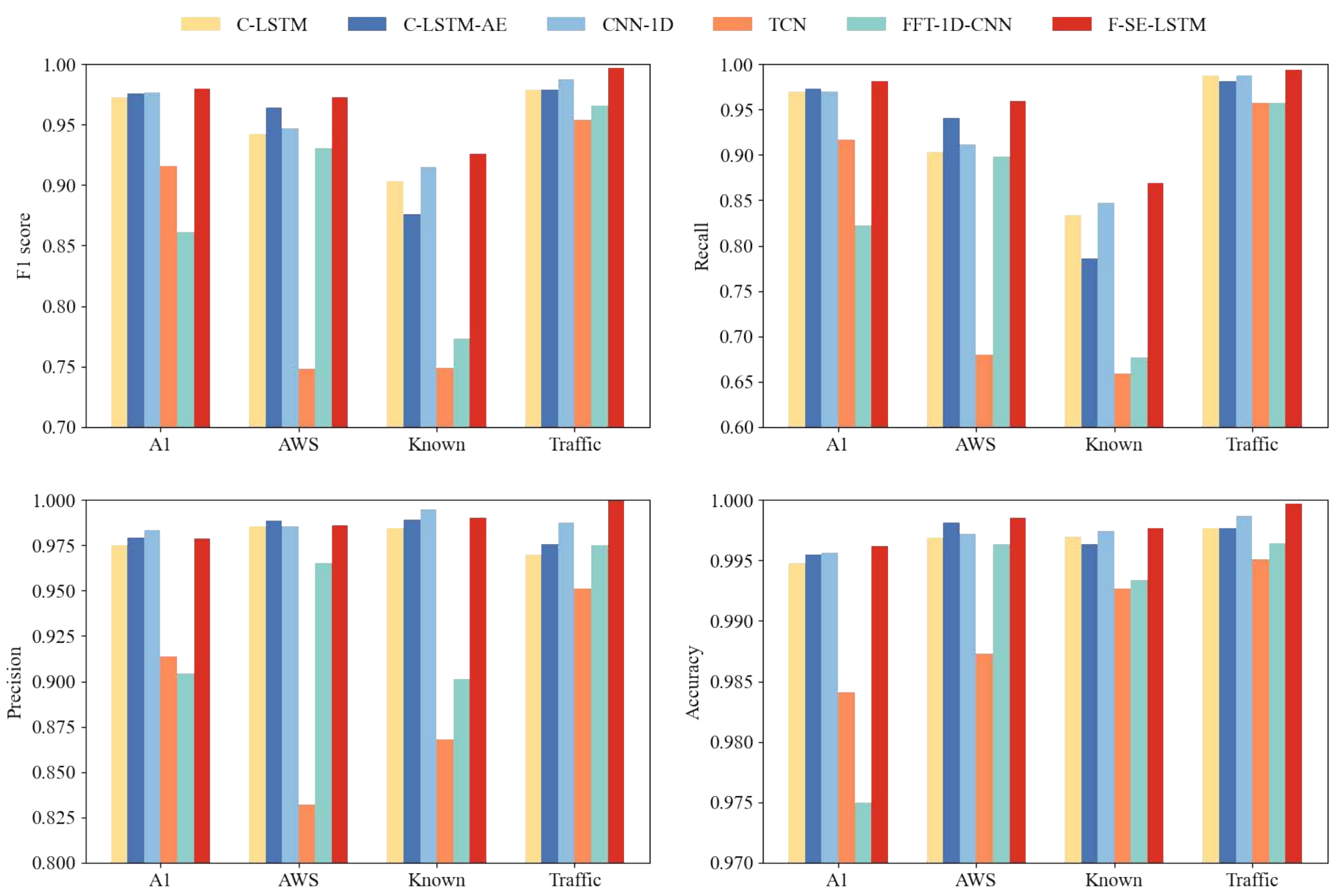}
    \caption{Experimental results of various methods.}
    \label{e5_f}
\end{figure*}

\begin{table*}[htbp]
    \centering
    \caption{Comparison of various methods on F1 scores in each dataset}
    \vspace{2px}
    % \scalebox{0.8}{
    \begin{tabular}{l|cccc}
        \toprule
        \textbf{Methods}                  & \textbf{A1}     & \textbf{AWS}    & \textbf{Known}  & \textbf{Traffic} \\
        \midrule
        C-LSTM \cite{kim2018web}          & 0.9725          & 0.9425          & 0.9031          & 0.9787           \\
        C-LSTM-AE \cite{yin2020anomaly}   & 0.9760          & 0.9642          & 0.8759          & 0.9786           \\
        CNN-1D \cite{ullah2021design}     & 0.9765          & 0.9469          & 0.9151          & 0.9877           \\
        TCN    \cite{neupane2022temporal} & 0.9155          & 0.7485          & 0.7494          & 0.9541           \\
        FFT-1D-CNN \cite{rahimi2020deep}  & 0.8612          & 0.9304          & 0.7731          & 0.9659           \\
        F-SE-LSTM [proposed]              & \textbf{0.9799} & \textbf{0.9728} & \textbf{0.9256} & \textbf{0.9969}  \\
        \bottomrule
    \end{tabular}
    % }
    % \begin{tablenotes}
    %     \footnotesize
    %     \item The time column represents how long the anomaly detection method was trained on the A1 dataset.
    % \end{tablenotes}
    \label{e5_t}
\end{table*}

\subsubsection{Comparison between our method and state-of-the-art methods}
We compare the proposed method with existing state-of-the-art methods.
For the fairness of the experiment, the sample sequence length of all these methods is uniformly set to 60. Fig. \ref{e5_f} shows the evaluation metrics of F1, recall, precision and accuracy of various methods on four datasets. Table \ref{e5_t} shows the performance comparison of various methods on multiple datasets. It can be seen that the proposed method outperforms other anomaly detection methods in terms of F1, recall and precision on all datasets, and also has higher precision than other methods.

Table \ref{e5_final_t} gives the comparison between our method and other methods from multiple dimensions such as performance, number of model parameters and training time: the average score of the proposed method in all evaluation metrics exceeds other anomaly detection methods, and it also outperforms most other methods in terms of training efficiency and storage space occupation.

Both C-LSTM and C-LSTM-AE methods are based on time series time-domain data, using neural networks such as CNN and LSTM to extract corresponding features according to the characteristics of time-domain data, and achieve high accuracy in anomaly detection.
The CNN-1D method has a higher accuracy in anomaly detection, but due to the large scale of the constructed model, a large storage space is required, and the training speed of the model is relatively slow.
TCN is a temporal convolutional neural network designed specifically for time series data in recent years. It mainly uses convolution operation to capture the time dependence of each moment of the sequence, and has good anomaly detection ability.
In contrast, our method not only excels in anomaly detection, but also in terms of training efficiency and storage space consumption.

FFT-1D-CNN is a frequency-domain based anomaly detection method that combines FFT with one-dimensional CNN. This method uses frequency data and can achieve good anomaly detection results using only a model with a very small amount of parameters. Although this method has a high model training efficiency and a small storage space, it does not consider the relationship between frequencies in the same time period and the relationship between frequencies in different periods, resulting in no further improvement in the method's anomaly detection ability.
In contrast, our method constructs a frequency matrix that can better reflect the characteristics of the time series, and builds a model suitable for extracting the characteristics of the frequency matrix, which greatly improves the anomaly detection ability.

In summary, the proposed method is better suited to the problem of anomaly detection in time series than the existing state-of-the-art methods.

\begin{table*}[htbp]
    \centering
    \caption{Comparison of various methods on the average F1, recall, precision, accuracy, training time and number of model parameters}
    \vspace{2px}
    % \scalebox{0.8}{
    \begin{tabular}{l|cccc|c|c}
        \toprule
        \textbf{Methods}                  & \textbf{F1}     & \textbf{Recall} & \textbf{Precision} & \textbf{Accuracy} & \textbf{Time} & \textbf{\#Parameter} \\
        \midrule
        C-LSTM \cite{kim2018web}          & 0.9492          & 0.9237          & 0.9788             & 0.9966            & 398s          & 56354                \\
        C-LSTM-AE \cite{yin2020anomaly}   & 0.9487          & 0.9203          & 0.9832             & 0.9969            & 682s          & 210076               \\
        CNN-1D \cite{ullah2021design}     & 0.9566          & 0.9290          & 0.9879             & 0.9972            & 433s          & 742466               \\
        TCN    \cite{neupane2022temporal} & 0.8419          & 0.8034          & 0.8913             & 0.9898            & 708s          & 64298                \\
        FFT-1D-CNN \cite{rahimi2020deep}  & 0.8826          & 0.8384          & 0.9364             & 0.9903            & \textbf{193s} & \textbf{15944}       \\
        F-SE-LSTM [proposed]              & \textbf{0.9688} & \textbf{0.9510} & \textbf{0.9887}    & \textbf{0.9980}   & 379s          & 35654                \\
        \bottomrule
    \end{tabular}
    % }
    % \begin{tablenotes}
    %     \footnotesize
    %     \item The time column represents how long the anomaly detection method was trained on the A1 dataset.
    % \end{tablenotes}
    \label{e5_final_t}
\end{table*}

\section{Conclusion}

In our work, we propose an anomaly detection method for time series, termed F-SE-LSTM, from the perspective of the frequency domain:
we first utilize sliding windows and FFT to construct the frequency matrix, and then extract frequency-related information within and between time periods with SENet and LSTM respectively.
Comparative results with other methods that the frequency matrix constructed by our method exhibits superior discriminative ability compared to ordinary time domain and frequency domain data, as it fully leverages the information of the frequency matrix and demonstrates outstanding performance on multiple real datasets.
Compared with existing deep learning-based anomaly detection methods, our approach achieves higher average F1 score and average recall rates, surpassing the best methods by 1.2\% and 2.2\% respectively.
Furthermore, this method also outperforms most other existing methods in terms of training speed and number of model parameters.

In future work, we aim to integrate both time domain and frequency domain data, striving to develop a more robust time series anomaly detection method.

\section*{Acknowledgment}
This work was partially supported by
Natural Science Foundation of Guangdong Province, China (Grant No.2021A1515011314),
Guangdong-Macao Science and Technology Cooperation Project (Grant No.2022A050520013 \& No.0059/2021/AGJ),
National Natural Science Foundation of China (Grant No.92067108),
"Qing Lan Project" in Jiangsu universities,
XJTLU RDF-22-01-020 and Pearl River Talent Program.



\bibliographystyle{elsarticle-num}
\bibliography{mybibfile}

\end{document}